\documentclass{article}
\usepackage{colortbl}
\usepackage{xcolor}
\usepackage{amsmath}
\usepackage{amssymb}
\usepackage{mathtools}
\usepackage{PRIMEarxiv}
\usepackage{multirow}
\usepackage[utf8]{inputenc} 
\usepackage[T1]{fontenc}    
\usepackage{hyperref}       
\usepackage{url}            
\usepackage{booktabs}       
\usepackage{amsfonts}       
\usepackage{nicefrac}       
\usepackage{microtype}      
\usepackage{lipsum}
\usepackage{fancyhdr}       
\usepackage{graphicx}       
\graphicspath{{media/}}     
\usepackage{algorithmic}
\usepackage{algorithm}
\usepackage{amsthm}

\newtheorem{theorem}{Theorem}
\newtheorem{lemma}{Lemma}

\title{Biased Local SGD for Efficient Deep Learning on Heterogeneous
 Systems}

\author{
  Jihyun Lim \\
  Inha University\\
  \texttt{wlguslim@inha.edu} \\
   \And
  Junhyuk Jo \\
  Inha University\\
  \texttt{911whwnsgur@inha.edu} \\
  \And
  Chanhyeok Ko \\
  Mondrian AI\\
  \texttt{chanhyeok@mondrian.ai} \\
  \And
  Young Min Go \\
  Mondrian AI\\
  \texttt{ymgo@mondrian.ai} \\
  \And
  Jimin Hwa \\
   Mondrian AI\\
  \texttt{hwa@mondrian.ai} \\
  \And
  Sunwoo Lee \\
  Inha University\\
  \texttt{sunwool@inha.ac.kr} \\
}

\begin{document}
\maketitle

\begin{abstract}
Most parallel neural network training methods assume homogeneous computing resources.
For example, synchronous data-parallel SGD suffers from significant synchronization overhead under heterogeneous workloads, often forcing practitioners to rely only on the fastest devices (e.g., GPUs).
In this work, we study local SGD for efficient parallel training on heterogeneous systems.
We show that intentionally introducing bias in data sampling and model aggregation can effectively harmonize slower CPUs with faster GPUs.
Our extensive empirical results demonstrate that a carefully controlled bias significantly accelerates local SGD while achieving comparable or even higher accuracy than synchronous SGD under the same epoch budget.
For instance, our method trains ResNet20 on CIFAR-10 with 2 CPUs and 8 GPUs up to $32\times$ faster than synchronous SGD, with nearly identical accuracy.
These results provide practical insights into how to flexibly utilize diverse compute resources for deep learning.
\end{abstract}


\section{Introduction}

Most parallel neural network training methods assume homogeneous computing resources.
For example, synchronous data-parallel stochastic gradient descent (SGD), the most widely used parallel training strategy, suffers from significant synchronization overhead when workers process their assigned data at different speeds.
Local SGD with periodic averaging, a fundamental distributed optimization algorithm in federated learning, faces the same limitation.
Although it reduces communication costs by decreasing the frequency of model aggregations, it still incurs substantial synchronization overhead at each aggregation step.
As a result, on modern GPU servers, relatively slower CPU resources often remain underutilized during training, despite being powerful computing units.

Some prior works have proposed training strategies designed to exploit heterogeneous system resources.
BytePS dedicates CPUs to summation operations while GPUs compute gradients~\cite{jiang2020unified}.
DLB distributes each mini-batch to heterogeneous resources using different local batch sizes~\cite{yeDLB}.
DistDGL utilizes both CPUs and GPUs to train large-scale graph neural networks~\cite{zheng2020distdgl}.
FusionFlow is a load scheduler that uses CPUs to assist slower GPUs~\cite{kim2023fusionflow}.
Metis is another load-balancing method designed for parallel training on heterogeneous GPUs~\cite{um2024metis}.
While these previous works utilize CPUs for training, they assume conventional synchronous SGD as the optimizer, thereby overlooking the potential of alternative distributed optimizers such as local SGD.
Moreover, although some of these methods introduce bias into gradient approximation, they do not systematically investigate its impact.

In this study, we focus on the efficient utilization of heterogeneous compute resources for neural network training.
Going beyond federated learning, we investigate the applicability of local SGD in centralized multi-process environments with independently and identically distributed (IID) data.
Specifically, we propose a local SGD–based parallel training framework that minimizes synchronization overhead across heterogeneous resources while achieving accuracy comparable to synchronous SGD.
First, we design a heterogeneous local SGD algorithm that allows each process to maximize the number of local training steps between consecutive model aggregations without blocking others.
This unbalanced local SGD effectively eliminates synchronization overhead in heterogeneous systems but may slow convergence due to unequal update frequencies across processes.
To address this issue, we intentionally introduce bias into local SGD–based training.

A carefully controlled bias is a key principle of our framework, enabling reduced synchronization cost while maintaining fast training convergence.
First, we strengthen the contribution of fast processes by assigning more critical data samples to them.
Our framework employs deterministic loss-based top-k sampling for fast processes to improve the quality of their local updates.

Second, we introduce additional bias in model aggregation by assigning larger weights to local updates from fast processes.
This improves the quality of global updates even when local updates from slow processes are insufficiently accurate.
We show that these two independently introduced, system-aware biases enable more effective utilization of heterogeneous system resources. 

To evaluate the performance of the proposed system-aware biased local SGD framework, we perform a set of representative computer vision (CV) and natural language processing (NLP) benchmarks under simulated heterogeneous system settings.
This study focuses on parallel training on a single GPU node with up to 2 CPUs and 8 GPUs.
Empirical results show that our framework effectively eliminates synchronization time across heterogeneous resources with large performance gaps, while preserving model accuracy.
E.g., our framework trains ResNet20 for 100 epochs $32\times$ faster than synchronous data-parallel SGD using an AMD EPYC 2543 CPU and two NVIDIA RTX 4090 GPUs, without compromising accuracy.

Our main contributions are summarized as follows.
\begin{itemize}
    \item Our study explores a novel local SGD–based approach for efficient parallel neural network training on heterogeneous compute resources. Our extensive empirical study demonstrates that unbalanced local SGD is a suitable optimizer for deep learning in heterogeneous environments, owing to its efficiency and flexibility.
    \item We also show that the well-controlled bias injected into global model updates effectively accelerates unbalanced local SGD, thereby allowing it to exploit heterogeneous system resources without any loss of accuracy.
    \item To the best of our knowledge, this is the first study to investigate biased local SGD on heterogeneous system resources under independently and identically distributed (IID) data settings. Our theoretical analysis shows that the proposed biased local SGD still guarantees convergence for smooth non-convex problems.
\end{itemize}
\section {Background} \label{sec:back}

\textbf{Problem Definition} --
In this work, we consider non-convex and smooth optimization problems as follows.
\begin{align}
    \min_{x \in \mathbb{R}^d} F(x) := \frac{1}{n} \sum_{i=1}^n f(x, \xi_i), \label{eq:problem_main}
\end{align}
where $f(x)$ is the loss function, $x \in \mathbb{R}^d$ is the $d$-dimensional model parameter vector, $n$ is the number of training samples, and $\xi_i$ is the $i^{th}$ training sample.

\textbf{Local Stochastic Gradient Descent} --
Local stochastic gradient descent (SGD) is a variant of SGD designed for communication-efficient model training.
First, it replicates the model to all workers.
Then, each worker locally trains the received model using a numerical optimization algorithm such as mini-batch SGD.
After a certain number of updates, the local models are averaged globally.
These steps are called \textit{communication round}.
The update rule is written as follows.
\begin{align}
    x_{t+1} = \frac{1}{m}\sum_{i=1}^{m}x_{t,\tau - 1}^{i} = x_t -\eta \frac{1}{m}\sum_{i=1}^{m} \sum_{j=0}^{\tau-1} \nabla f(x_{t,j}^i, \xi_j^i),
\end{align}
where $\tau$ is the number of local updates within each communication round, $\xi_j^i$ is the $j^{th}$ training sample drawn by worker $i$, and $m$ is the number of workers. Local SGD repeatedly runs communication rounds until the global model converges.
In modern high-performance or large-scale cloud computing platforms, all individual workers can access the entire dataset thanks to parallel file systems.
As a result, the dataset is IID, and thus $\mathbb{E}_{\xi \sim D}[\nabla f(x^i)] = \nabla F(x^i), \ \forall i \in [m]$.

\textbf{Biased Gradient Approximation} --
To efficiently solve the minimization problem shown in (\ref{eq:problem_main}), optimizers such as SGD approximate the first-order gradient of $F(\cdot)$ using randomly sampled data.
Mini-batch version of SGD is typically used to make a practical trade-off between system efficiency and stochastic efficiency.
When the batch is sampled based on a uniform random distribution, the approximated gradient is unbiased such that $\mathbb{E}_{\xi \sim D}[\nabla f(x, \xi)] = \nabla F(x)$.
In this study, however, we break this convention such that $\mathbb{E}_{\xi \sim p}[\nabla f(x, \xi)] = \nabla F_p(x) \neq \nabla F(x)$ and explore how this bias affects the parallel neural network training.

\section{Related work}

\textbf{Local SGD under IID Settings} --
Some prior works provide a theoretical foundation for local SGD under IID settings~\cite{stich2018local,haddadpour2019local,khaled2020tighter}.
Recently, STL-SGD has been proposed, which accelerates local SGD by adaptively adjusting model aggregation period~\cite{shen2021stl}.
While these previous works consider local SGD under IID settings, they either focus on theoretical performance guarantees or do not consider the system efficiency aspects.

\textbf{Biased Gradient Approximation} --
Several prior works have explored the use of bias in deep learning.
\textit{Active bias} is a training strategy that introduces bias toward high-variance samples~\cite{chang2017active}.
\textit{DIHCL} is a curriculum learning strategy that assigns higher weights to \lq{}hard-to-learn\rq{} samples.
An importance sampling method has also been proposed that estimates each sample’s significance using an approximated gradient norm bound~\cite{katharopoulos2018not}.
More recently, a classification strategy based on the population stability index was introduced to address class-imbalanced problems~\cite{yu2022re}.
Although these methods effectively leverage different forms of bias to accelerate training, they do not take into account the available system resources.

\textbf{Deep Learning on Heterogeneous Systems} --
Several previous studies have proposed deep learning approaches tailored to exploit heterogeneous system resources.
\textit{BytePS} assigns CPUs exclusively to summation tasks, while GPUs are responsible for gradient computation~\cite{jiang2020unified}.
\textit{DLB} enhances synchronous SGD by dynamically adjusting batch sizes based on individual worker performance~\cite{yeDLB}.
\textit{DistDGL} uses CPUs and GPUs together to train large-scale graph neural networks efficiently.
\textit{FusionFlow} acts as a load scheduler, assigning auxiliary tasks to CPUs to support slower GPUs~\cite{kim2023fusionflow}.
\textit{Metis} is another load-balancing framework designed for distributed training across heterogeneous GPUs~\cite{um2024metis}.
Although these methods effectively exploit CPU resources to accelerate synchronous SGD, they tend to overlook the potential of alternative distributed optimizers, such as local SGD.
Moreover, while these methods inherently introduce bias into gradient approximation, they do not examine its impact.

All these prior works either focus on biased gradient approximation without considering the underlying system resources, or on parallel training schemes that overlook the inherent bias they introduce.
In this study, we explore how system-aware bias can be exploited to better utilize heterogeneous resources for fast and accurate deep learning.
\section{Method}
In this section, we first introduce a local SGD-based neural network training approach designed to utilize heterogeneous system resources.
We then present two bias injection techniques and describe how they accelerate the training.

\subsection {Local SGD on Heterogeneous Resources}

\begin{figure}[t]
\centering
\includegraphics[width=0.65\columnwidth]{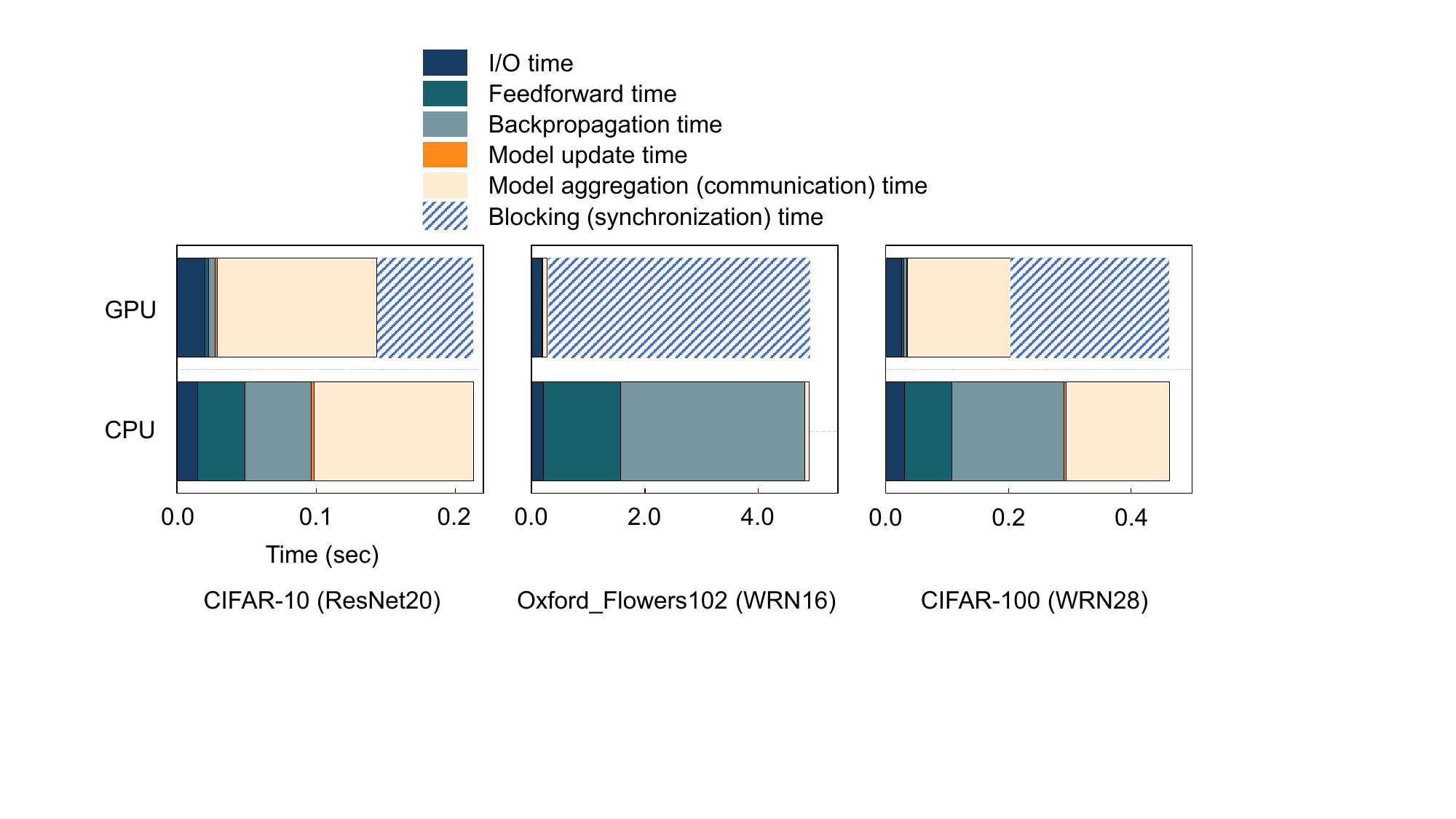}
\caption{
    Timing breakdown of a single training step, measured on an AMD EPYC 7543 CPU and two NVIDIA GeForce RTX 4090 GPUs. CPU training is an order of magnitude slower than GPU training. The resulting timing gap causes GPUs to remain idle, representing the synchronization cost.
}
\label{fig:timing}
\vspace{-0.3cm}
\end{figure}

\textbf {Motivation} --
We begin by discussing critical challenges in utilizing relatively slow compute resources, such as CPUs, for parallel neural network training.
The most common parallelization strategy is synchronous SGD with data-parallelism.
However, modern servers often contain heterogeneous compute resources, such as CPUs and GPUs with varying compute capabilities.
In such systems, synchronous SGD and balanced local SGD, where all workers perform the same number of local updates per communication round, incur high synchronization costs, as faster workers must wait for slower ones to reach the synchronization point. 

Figure~\ref{fig:timing} shows a timing breakdown of neural network training on a GPU server equipped with an AMD EPYC 7543 CPU and two NVIDIA GeForce RTX 4090 GPUs.
The reported timings correspond to the elapsed time for processing a single mini-batch of size 128.
Across all datasets, synchronous SGD consistently suffers from substantial GPU blocking time.
In particular, the Oxford Flowers 102 dataset involves relatively larger images than the CIFAR datasets, resulting in a noticeable computational efficiency gap between the CPU and GPU.
When local SGD is applied, the total number of model aggregations is reduced, thereby lowering communication costs; however, the blocking time of faster workers remains unchanged.
As the performance gap between compute resources widens, this overhead becomes increasingly significant.
This observation motivates us to explore a novel local SGD–based parallel training framework that explicitly addresses synchronization overhead.

\begin{figure*}[t]
\centering
\includegraphics[width=0.9\columnwidth]{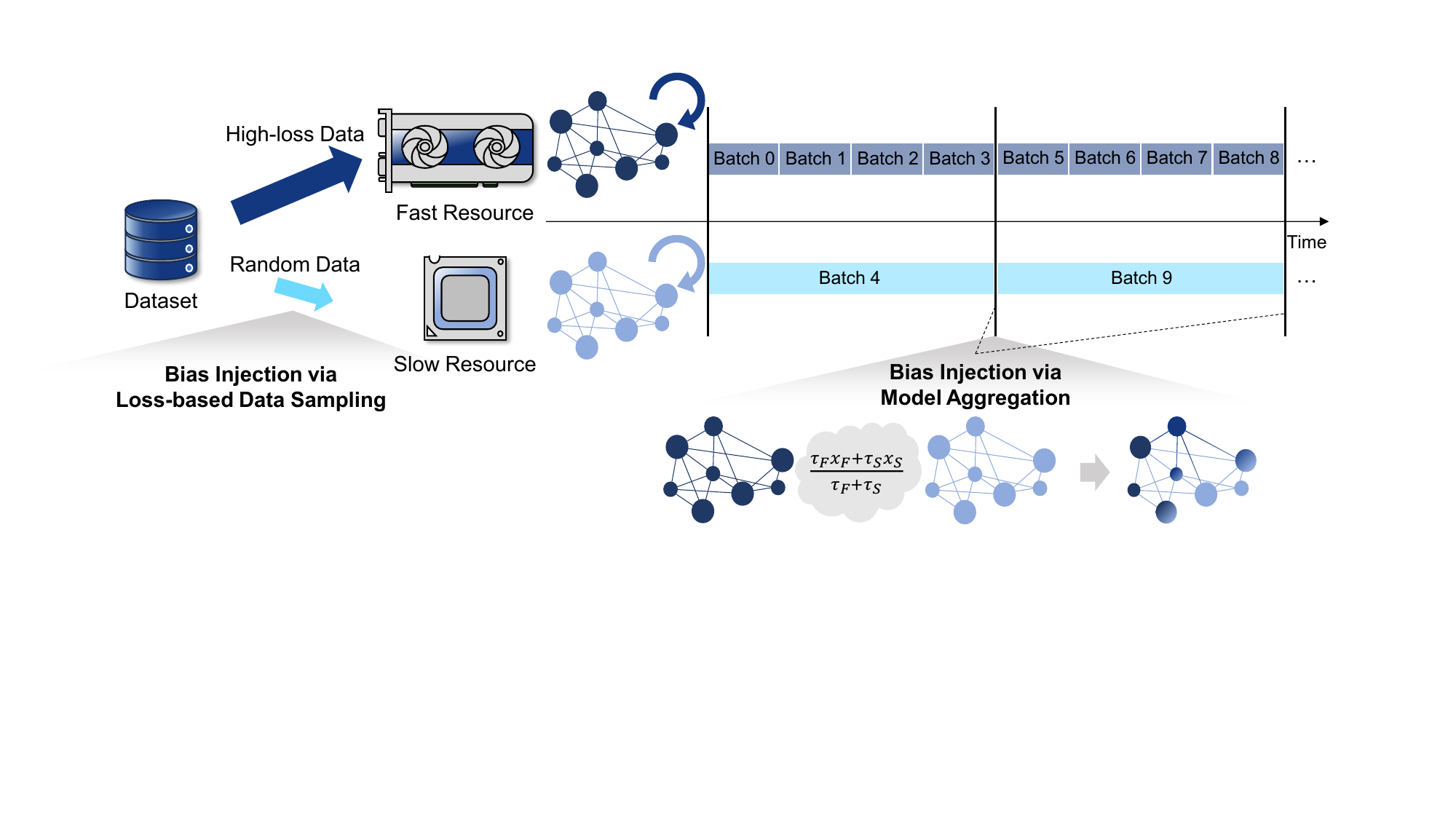}
\vspace{-0.5em}
\caption{
    Schematic illustration of system-aware biased local SGD framework. The $\tau_F$ and $\tau_S$ are the number of local updates per communication round on fast and slow resources, respectively. The $x_F$ and $x_S$ are the models locally trained on fast and slow resources (e.g., GPU and CPU), respectively.
}
\label{fig:schematic}
\vspace{-0.7em}
\end{figure*}

\textbf {System-aware Unbalanced Local SGD Framework} --
We propose a general local SGD framework for parallel neural network training.
To eliminate the synchronization cost caused by the performance gap across compute resources, we design a system-aware unbalanced local SGD.
Specifically, a worker utilizing faster resources performs more local updates than one using slower resources.
Figure~\ref{fig:schematic} shows a schematic illustration of our framework.

We define two notations: $\tau_S$ and $\tau_F$, representing the number of local updates per communication round on slow and fast resources, respectively.
For simplicity, we only consider two types of system resources: fast and slow.
However, generalizing this to a larger number of resource types is straightforward.
We also define a \textit{performance gap ratio}, $0 < \gamma \leq 1$, as the ratio of the processing time for a single mini-batch on a fast resource to that on a slow resource.
This $\gamma$ can be easily measured once on the target hardware platform and reused later.
Therefore, we assume that $\gamma$ is provided in advance.

Our framework first allows users to tune $\tau_F$ as with other typical hyperparameters.
After tuning $\tau_F$, we set $\tau_S = \gamma \tau_F$.
In this way, the synchronization cost (the blocking time of fast resources) can be eliminated by reducing the workload on slow resources, as illustrated in Figure~\ref{fig:schematic}.
As the performance gap between slow and fast resources increases, $\gamma$ decreases, which in turn results in fewer training iterations on the slow resources.

While this unbalanced local SGD approach eliminates the synchronization cost, it may slow down convergence due to the reduced number of updates on slow resources.
To address this issue, we propose introducing a controlled bias into local SGD training.
Our framework comprises two key bias-injection methods: (1) loss-based data sampling and (2) system-aware model aggregation, as described in the following subsections.

\subsection {Bias Injection via Data Sampling} \label{sec:sample}
One key component of our framework is system-aware biased data sampling.
The core idea is to allocate more important data to faster workers so that the model is more exposed to critical information.
Our approach employs a loss-based top-$k$ selection strategy.
This deterministic sampling ensures that challenging data are assigned to faster workers, thereby accelerating convergence.

At the beginning of each epoch, fast and slow workers perform data sampling as follows.
\begin{itemize}
    \item \textbf{Data Allocation for Fast Workers}: At each epoch, $P_F$ fast workers consume $\tau_F P_F$ mini-batches while $P_S$ slow workers consume $\tau_S P_S$ batches. Thus, given $N$ total samples, the fast workers are allocated $N_F := \frac{\tau_F P_F}{\tau_F P_F + \tau_S P_S} N$ samples. First, $\lambda N_F$ samples with the highest loss values are selected, where $0 < \lambda \leq 1$ is a hyper-parameter which determines the ratio of high-loss samples. Then, $(1 - \lambda) N_F$ samples are randomly selected from the rest of data. Finally, these $N_F$ samples are evenly distributed to $P_F$ fast workers.
    \item \textbf{Data Allocation for Slow Workers}: Given $N$ samples, the $P_S$ slow workers are assigned $N_S := \frac{\tau_S P_S}{\tau_F P_F + \tau_S P_S} N$ samples per epoch. First, $N_S$ samples are randomly sampled from the total $N$ samples without replacement. These samples are then evenly distributed among the $P_S$ slow workers.
\end{itemize}

This biased data sampling strategy has two desirable characteristics.
First, the fast workers primarily focus on samples with high loss values, thereby introducing bias into their gradient approximations toward challenging samples.
Second, the slow workers perform uniform random sampling independently of the fast workers. That is, while they do not introduce bias, they may still process high-loss samples.
Consequently, the model is expected to focus more on challenging data by allocating a larger portion of the iteration budget to learning from them.

When implementing the above biased data sampling, we let a single process perform all the sampling and then broadcast the resulting indices to all processes.
This design choice minimizes implementation complexity by centralizing the sampling workload, while incurring only minor overhead.
For example, when $N = 100{,}000$, the total size of the indices is only about 390 KB, so collecting $N$ loss values and executing the two steps above introduces negligible additional computation and communication overhead.

\begin{algorithm}[t]
\caption{System-aware biased local SGD framework.}
\label{alg:framework}
\begin{algorithmic}[1]
\small
\REQUIRE $\gamma$: the ratio of the mini-batch processing time on faster resources to that of slower ones, $\lambda$: the ratio of high-loss samples selected from the dataset assigned to fast resources, $\tau_F$: the number of local updates per round on fast resources, $\tau_S$: the number of local updates per round on slow resources,  $P_S$: the number of slow workers,  $P_F$: the number of fast workers.

\STATE {$\tau_S \leftarrow \gamma \tau_F$}
\FOR{$t = 1 \to E$}
    \STATE Select $\frac{\tau_F P_F}{\tau_S P_S + \tau_F P_F} \lambda N$ high-loss samples.
    \STATE Fill the remainder of the fast workers' allocated data by randomly sampling from the unselected samples.
    \STATE Evenly distribute the selected samples among all $P_F$ fast workers.
    \STATE Allocate $\frac{\tau_S P_S}{\tau_S P_S + \tau_F P_F} N$ samples evenly among the $P_S$ slow workers.

    \WHILE{the whole dataset is not covered yet}
        \IF {worker $i$ is a slow worker}
            \STATE {Locally update $x_t$ for $\tau_S$ iterations.}
        \ENDIF
        \IF {worker $i$ is a fast worker}
            \STATE {Locally update $x_t$ for $\tau_F$ iterations.}
        \ENDIF
        \STATE {Record all individual loss of processed data.}
        \STATE {$x_{t+1} \leftarrow$ Aggregate the local models, $x_t^i$, using Eq. (\ref{eq:aggregate}).}
    \ENDWHILE
\ENDFOR
\STATE \textbf{Return:} $x_{E}$
\end{algorithmic}
\end{algorithm}

\subsection {Bias Injection via Model Aggregation} \label{sec:aggregation}
Another key component of our framework is system-aware model aggregation.
Under independent and identically distributed (IID) settings, local SGD aggregates the locally trained models as follows.
\begin{align}
    x_{t+1} = \frac{1}{m} \sum_{i=1}^{m} x^i_{t},
\end{align}
where $m$ is the number of workers and $x^i_t$ is the local model trained by worker $i$ at round $t$.
This balanced averaging ensures that the global update is unbiased and thus minimizes the loss $\frac{1}{n}\sum_{i=1}^{n} f(x, \xi_i)$, where $n$ is the number of training samples, $\xi_i$ is the $i$‑th training sample, and $f(\cdot, \cdot)$ denotes the loss function.
In contrast, our framework breaks this convention by assigning greater weight to more important local models.

We propose aggregating $m$ locally trained models using the following equation.
\begin{align}
    x_{t+1} = \sum_{i=1}^{m} \frac{\tau_i}{\sum_{i=1}^m \tau_i} x^i_t, \label{eq:aggregate}
\end{align}
where $\tau_i$ is the number of local steps performed by worker $i$.
This aggregation method introduces bias toward local models that perform more updates.
The key idea behind this design is to let the more extensively trained models guide the global update.
By assigning greater weight to local models from faster resources, the global model becomes biased in their favor.


\subsection {System-aware Biased Local SGD Framework}
Algorithm~\ref{alg:framework} presents our biased local SGD framework.
At each round, fast workers select their high-loss data and locally train the model for $\tau_F$ iterations.
Likewise, slow workers randomly sample data and then train the model for $\tau_S$ iterations.
By parallelizing the loop at line 7, our framework effectively utilizes heterogeneous system resources.
Finally, all the locally trained models are aggregated using ($\ref{eq:aggregate}$) at the end of each communication round.


\textbf{Potential Limitations} --
Our proposed method has the following limitations.
First, it requires slightly more memory than unbiased local SGD because the loss values of all individual training samples must be stored.
However, given the size of modern neural networks, the overhead of storing a single floating-point number per sample is negligible.
Second, the method involves additional computation to sort the loss values and select the highest-loss samples.
Since this biased data selection is performed only once after consuming all the assigned data, the computational overhead remains minimal.

\subsection{Convergence Analysis}\label{sec:theory}
The proposed system-aware biased local SGD comes with theoretical convergence guarantees.
The problem definition can be found in Section~\ref{sec:back}.
To analyze the impact of introducing bias into local SGD, we define the expected stochastic gradient based on a certain probability distribution $p$ as follows.
\begin{align}
    \mathbb{E}_{\xi\sim p}\left[\nabla f(x, \xi)\right] = \nabla F_p(x).
\end{align}
If $p_i, i\in [n]$ is $\frac{1}{n}$ for all data samples, this corresponds to unbiased uniform random sampling: $\nabla F_p(x) = \nabla F(x)$.
Otherwise, it corresponds to biased sampling that yields:
$\nabla F_p(x) \neq \nabla F(x)$.

We also define \lq{}global update\rq{} aggregated using a certain weight factor $w_i$, $\bar\Delta_t$ and its expectation as follows.
\begin{align}
    \bar\Delta_t &= \sum_{i=1}^{m}\sum_{j=0}^{\tau - 1}w_i g_{t,j}^i := \sum_{i=1}^{m}\sum_{j=0}^{\tau - 1} w_i \nabla f(x_{t,j}^i, \xi) \nonumber \\
    \mathbb{E}_{\xi \sim p} \left[ \bar \Delta_t \right] &= \mathbb{E}_{\xi \sim p} \left[ \sum_{i=1}^{m}\sum_{j=0}^{\tau - 1} w_i g_{t,j}^i \right] \nonumber \\
    &\qquad := \sum_{i=1}^{m} \sum_{j=0}^{\tau - 1} w_i \nabla F_p(x_{t,j}^i), \nonumber
\end{align}
where $m$ is the number of workers and $\tau$ is the number of local update steps.
The $w_i$ is aggregation weight factor which satisfies $\sum_{i=1}^m w_i = 1$.
This definition allows us to analyze the impact of biased model aggregation discussed in Section~\ref{sec:aggregation}.

Our analysis is based on the following assumptions.

\noindent
\textbf{Assumption 1.} \textit{(Continuity) There exists a constant $\mathcal{L}>0$, such that $\| \nabla F(x) - \nabla F(y) \| \leq \mathcal{L} \| x-y \|, \forall x,y \in \mathbb{R}^d$.}

\noindent
\textbf{Assumption 2.} \textit{(Biased gradients) The gradient estimator is biased based on a probability distribution $p$ such that $\mathbb{E}_{\xi \sim p}[\nabla f(x, \xi)] = \nabla F_p(x) \neq \nabla F(x)$.}

\noindent
\textbf{Assumption 3.} \textit{(Bounded variance) There exists a constant $\sigma > 0$, such that the stochastic gradient variance is bounded by $\mathbb{E}_{\xi \sim p}[\| \nabla f(x, \xi) - \nabla F_p(x) \|]^2 \leq \sigma_p^2$.}

Then, we analyze the convergence properties of biased local SGD in the IID case.

\noindent
\textbf{Lemma 1.}
\textit{(Fixed learning rate) Under assumption 1 $\sim$ 3, if the learning rate $\eta \leq \frac{1}{\mathcal{L}\tau}$, we have}
\begin{align}
\mathbb{E}_p\left[ F(x_{t+1}) \right] & \leq \mathbb{E}_p\left[ F(x_t) \right] \nonumber \\
&\qquad -\frac{\eta}{2} \mathbb{E}_p\left[ \left\| \nabla F(x_t) \right\|^2 \right] + \frac{\mathcal{L}\eta^2 \tau^2}{2} \sigma_p^2. \nonumber
\end{align}
Based on this bound, we derive the bound of (\ref{eq:problem})'s full gradient as follows.

\noindent
\textbf{Theorem 1.}
\textit{Under assumption $1 \sim 3$, if the learning rate $\eta \leq \frac{1}{\mathcal{L}\tau}$ and $\tau \geq 1$, the average magnitude of gradient becomes}
\begin{align}
    \frac{1}{T}\sum_{t=1}^{T} \mathbb{E}_p \left[ \left\| \nabla F(x_{t}) \right\|^2 \right] &\leq \frac{2}{\eta T} \left( \mathbb{E}_p \left[ F(x_0) \right] - \mathbb{E}_p \left[ F(x_{T}) \right] \right) \nonumber \\
    &\qquad + \mathcal{L}\eta \tau^2 \sigma_p^2. \label{eq:theorem}
\end{align}

\noindent
\textbf{Remark 1.}
\textit{The above results demonstrate that, under the IID condition, our proposed biased local SGD guarantees convergence regardless of how many workers participate in the training.
In addition, the right-most variance term in (\ref{eq:theorem}) becomes larger as $\tau$ increases.
Therefore, users can make a practical trade-off between the convergence rate (statistical efficiency) and the total number of communications (system efficiency) by carefully tuning $\tau$.
}

\noindent
\textbf{Remark 2.}
\textit{One intriguing observation is that local SGD achieves guaranteed convergence with a sufficiently small learning rate, even under biased data sampling and model aggregation.
Specifically, the convergence bound is independent of the data sampling distribution $p$ and the model aggregation weight factor $w$.
That is, for any choice of $p$ and $w$, the biased local SGD guarantees convergence under IID settings.
This result indicates that the proposed method can be safely applied in real-world deep learning applications, improving system utilization without compromising the convergence guarantee.
}

\section{Experiments}

\subsection {Experimental Settings}

\textbf{Benchmark Settings} -- 
We evaluate our system-aware biased local SGD framework on four computer vision benchmarks, CIFAR-10~\cite{krizhevsky2009learning}, CIFAR-100, Oxford Flowers102~\cite{nilsback2008automated}, and Tiny ImageNet~\cite{tiny-imagenet}, and one NLP benchmark, AG News~\cite{zhang2015character}.
We use ResNet20~\cite{he2016deep}, Wide-ResNet28~\cite{zagoruyko2016wide}, Wide-ResNet16, ViT~\cite{dosovitskiy2020image}, and DistilBERT~\cite{sanh2019distilbert}, respectively.
ViT is pretrained on ImageNet-1k~\cite{deng2009imagenet}, and DistilBERT on BookCorpus~\cite{bandy2021addressing} and Wikipedia.
All results are averaged over at least three independent runs.

\textbf{System Settings} --
We used a GPU server equipped with two NVIDIA RTX 4090 GPUs and one AMD EPYC 7543 CPU. Our experiments include the \textit{1-CPU, 1-GPU} and \textit{2-CPU, 8-GPU} settings, which reflect typical heterogeneous configurations within a single GPU server node.
To vary the degree of CPU–GPU heterogeneity, we adjust $\tau_S$, the number of local updates per communication round performed by the slower resource.
E.g., when $\tau_F = 32$, if the slower resource has half the compute power of the faster one, we set $\tau_S = 16$.
The fastest compute resource in our experimental setup is the NVIDIA RTX 4090, which takes 0.1737 seconds to process a single mini-batch of Oxford Flowers102 using Wide-ResNet16.
In contrast, the slowest resource, the AMD EPYC 7543, requires 3.8906 seconds for the same workload.
This results in a performance ratio of approximately $22$ times.
To reflect this performance gap, we set $\tau_F = 32$ and $\tau_S = 1$. In our empirical study, we simulate three heterogeneous settings: $(\tau_F, \tau_S) = (32, 16)$, $(32, 4)$, and $(32, 1)$. In the main manuscript, we fixed $\tau_F=32$ in all the experiments.
See the Appendix for all hyper-parameter settings.

\begin{table}[h]
\footnotesize
\centering
\caption{
   Performance comparison among various optimizers on 2-CPU and 8-GPU. When the epoch budget is the same, biased local SGD shows a slight decline in accuracy while dramatically reducing the training time.
}
\begin{tabular}{lcccrc} \toprule
\multirow{2}{*}{Dataset} & \multirow{2}{*}{Algorithm} & \multirow{2}{*}{Epochs} & Comm. & Wall-clock & \multirow{2}{*}{Accuracy} \\ 
& & & per epoch &Time (sec)  & \\ \midrule
\multirow{7}{*}{CIFAR-10}& Synchronous SGD + Data-Parallelism & 100  &  128 & 1298.39 & $90.91\pm 0.1\%$ \\ 
& Snychronous SGD + Data-Parallelism (8-GPU only) & 100 & 160& 1234.75 & $91.28
\pm 0.3 \%$\\
\multirow{5}{*}{(ResNet20)}& Balanced Local SGD & 100 &  4& 548.04 & $88.93\pm 0.2\%$ \\ 
                            & BytePS & 100 &160& 1279.43& $90.99\pm0.3  \%$ \\ 
                           & DLB & 100 &128  &987.80 &$86.89\pm0.2 \%$ \\
                      
& \multicolumn{1}{>{\columncolor{gray!10}}c}{System-aware Biased Local SGD(Proposed)}
& \multicolumn{1}{>{\columncolor{gray!10}}c}{100}
& \multirow{3}{*}{\cellcolor{gray!10}}{4}
& \multicolumn{1}{>{\columncolor{gray!10}}r}{40.57}
& \multicolumn{1}{>{\columncolor{gray!10}}c}{$\mathbf{89.16}\pm 0.1\%$} \\
& \multicolumn{1}{>{\columncolor{gray!10}}c}{System-aware Biased Local SGD(Proposed)}
& \multicolumn{1}{>{\columncolor{gray!10}}c}{200}
& \multicolumn{1}{>{\columncolor{gray!10}}c}{4}   
& \multicolumn{1}{>{\columncolor{gray!10}}r}{81.15}
& \multicolumn{1}{>{\columncolor{gray!10}}c}{$\mathbf{91.26}\pm 0.1\%$} \\

& \multicolumn{1}{>{\columncolor{gray!10}}c}{System-aware Biased Local SGD(Proposed)}
& \multicolumn{1}{>{\columncolor{gray!10}}c}{300}
& \multicolumn{1}{>{\columncolor{gray!10}}c}{4}   
& \multicolumn{1}{>{\columncolor{gray!10}}r}{121.72}
& \multicolumn{1}{>{\columncolor{gray!10}}c}{$\mathbf{91.95}\pm 0.1\%$} \\

                            \midrule
\multirow{7}{*}{Oxford}& Synchronous SGD + Data-Parallelism & 300  & 64  &34659.58 & $84.75\pm 0.9\%$ \\ 
& Snychronous SGD + Data-Parallelism (8-GPU only) & 300 & 80 &2866.08 & $85.88\pm 0.1 \%$\\
& Balanced Local SGD & 300 & 2 & 33497.26& $69.51\pm 1.6\%$ \\ 
                        & BytePS  & 300 & 80& 3091.76& $86.67\pm0.7  \%$ \\ 
                         Flowers 102  & DLB & 300 &64  &2292.87 &$76.52\pm0.9 \%$ \\
                         (WRN16)   & \multicolumn{1}{>{\columncolor{gray!10}}c}{System-aware Biased Local SGD(Proposed)}
                        & \multicolumn{1}{>{\columncolor{gray!10}}c}{300}
                        & \multicolumn{1}{>{\columncolor{gray!10}}c}{2}
                        & \multicolumn{1}{>{\columncolor{gray!10}}r}{1083.11}
                        & \multicolumn{1}{>{\columncolor{gray!10}}c}{$\mathbf{73.09}\pm 1.4\%$} \\
                        & \multicolumn{1}{>{\columncolor{gray!10}}c}{System-aware Biased Local SGD(Proposed)}
                        & \multicolumn{1}{>{\columncolor{gray!10}}c}{600}
                        & \multicolumn{1}{>{\columncolor{gray!10}}c}{2}
                        & \multicolumn{1}{>{\columncolor{gray!10}}r}{2166.22}
                        & \multicolumn{1}{>{\columncolor{gray!10}}c}{$\mathbf{85.29}\pm 0.9\%$} \\
                        & \multicolumn{1}{>{\columncolor{gray!10}}c}{System-aware Biased Local SGD(Proposed)}
                        & \multicolumn{1}{>{\columncolor{gray!10}}c}{900}
                        & \multicolumn{1}{>{\columncolor{gray!10}}c}{2}
                        & \multicolumn{1}{>{\columncolor{gray!10}}r}{3249.34}
                        & \multicolumn{1}{>{\columncolor{gray!10}}c}{$\mathbf{88.04}\pm 0.1\%$} \\
                            \midrule
\multirow{7}{*}{CIFAR-100}& Synchronous SGD + Data-Parallelism & 200  & 128 &80602.47 & $80.20\pm0.2 \%$ \\ 
& Snychronous SGD + Data-Parallelism (8-GPU only) & 200 & 160 & 48180.65 & $80.26\pm0.1  \%$\\
\multirow{5}{*}{(WRN28)}& Balanced Local SGD & 200 & 4 & 43952.17 & $78.24\pm0.1 \%$ \\ 
                         & BytePS  & 200 & 160 & 130155.6& $80.07\pm 0.1 \%$ \\ 
                         & DLB & 200 & 128 &38544.52 &$76.42\pm 0.1\%$ \\
                         & \multicolumn{1}{>{\columncolor{gray!10}}c}{System-aware Biased Local SGD(Proposed)}
                        & \multicolumn{1}{>{\columncolor{gray!10}}c}{200}
                        & \multicolumn{1}{>{\columncolor{gray!10}}c}{4}
                        & \multicolumn{1}{>{\columncolor{gray!10}}r}{2518.83}
                        & \multicolumn{1}{>{\columncolor{gray!10}}c}{$\mathbf{77.97}\pm 0.4\%$} \\
                        & \multicolumn{1}{>{\columncolor{gray!10}}c}{System-aware Biased Local SGD(Proposed)}
                        & \multicolumn{1}{>{\columncolor{gray!10}}c}{400}
                        & \multicolumn{1}{>{\columncolor{gray!10}}c}{4}
                        & \multicolumn{1}{>{\columncolor{gray!10}}r}{5037.66}
                        & \multicolumn{1}{>{\columncolor{gray!10}}c}{$\mathbf{80.03}\pm 0.1\%$} \\
                        & \multicolumn{1}{>{\columncolor{gray!10}}c}{System-aware Biased Local SGD(Proposed)}
                        & \multicolumn{1}{>{\columncolor{gray!10}}c}{600}
                        & \multicolumn{1}{>{\columncolor{gray!10}}c}{4}
                        & \multicolumn{1}{>{\columncolor{gray!10}}r}{7556.48}
                        & \multicolumn{1}{>{\columncolor{gray!10}}c}{$\mathbf{80.58}\pm 0.3\%$} \\
                            \bottomrule
\end{tabular}
\label{tab:compare}
\end{table}

\begin{figure}[h]
\centering
\includegraphics[width=0.8\columnwidth]{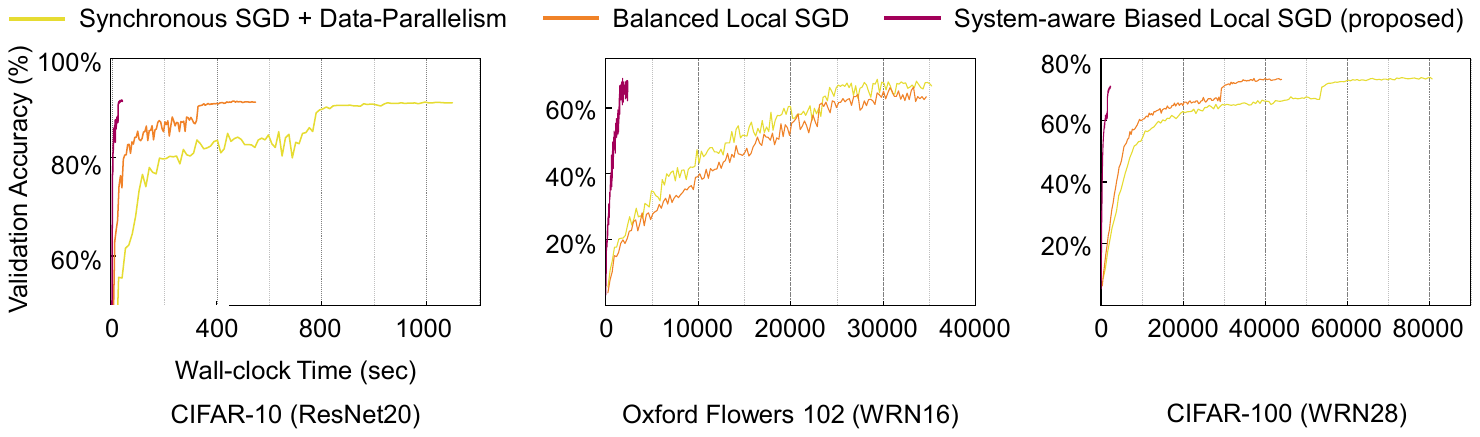}
\vspace{-0.5em}
\caption{
    Learning curve comparison. Thanks to reduced communication and synchronization costs, our proposed method completes the same number of training rounds much faster than synchronous SGD and balanced local SGD.
}
\label{fig:curves}
\end{figure}

\subsection {Comparative Study}
\textbf{Performance Comparison under a Fixed Epoch Budget} -- 
Table~\ref{tab:compare} compares various system-aware training algorithms in terms of model accuracy and wall-clock time for a fixed epoch budget.
Balanced local SGD is a conventional local SGD scheme in which all processes perform the same number of local updates.
BytePS~\cite{jiang2020unified} is a parallel training framework that offloads relatively simple computations, such as vector summations, to CPUs.
DLB~\cite{yeDLB} adaptively adjusts the local batch size based on the image throughput observed in the previous epoch.

First, balanced local SGD significantly reduces communication frequency, resulting in shorter training time than conventional synchronous SGD.
However, as shown in Figure~\ref{fig:timing}, it still incurs synchronization overhead.
Similarly, while BytePS and DLB enable the use of CPUs for training, they still suffer from significant synchronization costs.
Our proposed method eliminates this blocking time by adjusting $\tau_S$ in a system-aware manner, thereby removing synchronization overhead.
Consequently, it completes training faster than balanced local SGD.


\begin{table}[t]
\footnotesize
\centering
\caption{Performance comparison of GPU-only and heterogeneous CPU--GPU training settings.}
\begin{tabular}{llcc} \toprule
\multirow{2}{*}{Dataset} & \multirow{2}{*}{Algorithm} & Wall-clock & \multirow{2}{*}{Accuracy} \\
& &  Time (sec) & \\ \midrule
CIFAR-10& 8GPU-only  &  38.14 & $88.75\pm 0.1\%$\\ 
(ResNet20)&  2CPU–8GPU   & 40.57 & $\mathbf{89.16}\pm 0.1\%$\\  \midrule
Oxford &\multirow{2}{*}{8GPU-only}  &\multirow{2}{*}{1077.56}  & \multirow{2}{*}{$69.22\pm 0.8\%$}\\ 
Flowers102& \multirow{2}{*}{2CPU–8GPU} & \multirow{2}{*}{1083.11}& \multirow{2}{*}{$\mathbf{73.09}\pm 1.4\%$} \\
(WRN16)& & & \\  \midrule
CIFAR-100& 8GPU-only  & 2470.26 & $77.19\pm 0.5\%$\\ 
(WRN28)& 2CPU–8GPU   & 2518.83 &  $\mathbf{77.97}\pm 0.4\%$\\ \bottomrule
\end{tabular}
\label{tab:gpu_vs_cpu_gpu}
\end{table}

\begin{table}[t]
\caption{
    [\textbf{2-CPU and 8-GPU}] Performance comparison between the unbiased local SGD and the proposed biased local SGD. 
}\label{tab:c2g8}
\footnotesize
\centering
\begin{tabular}{llc>{\columncolor{gray!10}}c} \toprule
\multirow{2}{*}{Dataset} & \multirow{2}{*}{($\tau_F$, $\tau_S$)} & \multirow{2}{*}{Unbiased} & Biased \\ 
&&& (proposed) \\ \midrule
\multirow{2}{*}{CIFAR-10} & (32,16) & $88.83\pm 0.3\%$ & $\mathbf{89.71}\pm 0.2\%$ \\ 
\multirow{2}{*}{(ResNet20)} & (32,4)& $ 88.09\pm 0.1\%$ & $\mathbf{89.47}\pm 0.1\%$ \\ 
& (32,1)& $87.98 \pm 0.2\%$ & $\mathbf{89.16}\pm0.1 \%$ \\  \midrule
Oxford & (32,16) & $ 69.26\pm 1.3\%$ & $\mathbf{72.94}\pm 0.9\%$ \\ 
Flowers102 & (32,4) & $ 66.57\pm1.8 \%$ & $\mathbf{73.68}\pm 0.1\%$ \\ 
(WRN16) & (32,1) & $66.08 \pm1.0 \%$ & $\mathbf{73.09}\pm 1.4\%$ \\  \midrule
\multirow{2}{*}{CIFAR-100} & (32,16)& $ 77.75\pm0.6 \%$ & $\mathbf{78.28}\pm0.2 \%$ \\ 
\multirow{2}{*}{(WRN28)}& (32,4)& $ 77.33\pm 0.9\%$ & $\mathbf{78.44}\pm0.1 \%$ \\ 
& (32,1) & $77.01 \pm 0.2\%$ & $\mathbf{77.97}\pm 0.4\%$ \\ \midrule
Tiny & (32,16) & $ 90.00\pm 0.1\%$ & $\mathbf{90.52}\pm0.2 \%$ \\ 
ImageNet & (32,4) & $ 89.90\pm0.2 \%$ & $\mathbf{90.41}\pm0.1 \%$ \\ 
(ViT) & (32,1) & $ 89.12\pm 0.4\%$ & $\mathbf{90.43}\pm0.1 \%$ \\ \midrule
\multirow{2}{*}{AG News} & (32,16) & $93.38 \pm0.1 \%$ & $\mathbf{93.66}\pm 0.1\%$ \\ 
\multirow{2}{*}{(DistilBERT)} & (32,4) & $93.22 \pm0.1 \%$ & $\mathbf{93.57}\pm0.1 \%$ \\ 
& (32,1) & $93.17 \pm0.1 \%$ & $\mathbf{93.51}\pm0.1 \%$ \\ 
\bottomrule
\end{tabular}
\vspace{-1em}
\end{table}

Thanks to the dramatically reduced wall-clock time, one can increase the epoch budget without increasing the total elapsed time.
In Table~\ref{tab:compare}, we present the accuracy and training time of our method after training for $200\%$ and $300\%$ of the baseline number of epochs.
By running extra epochs, the elapsed time remains considerably shorter than that of other methods, while the accuracy becomes even higher than that of synchronous SGD.
This result clearly demonstrates the key benefit of our proposed biased local SGD on heterogeneous compute resources.

\noindent
\textbf{Effective Weak Scaling} --
We validate the key idea of leveraging \lq{}slow\rq{} resources by analyzing weak-scaling performance.
Table~\ref{tab:gpu_vs_cpu_gpu} presents accuracy comparisons between an 8-GPU-only configuration and a hybrid 2-CPU + 8-GPU configuration.
For a fixed wall-clock time, accuracy is consistently improved when CPUs participate in training across all benchmarks.
Although the observed speedup is far from ideal due to the capacity gap between resources, these weak-scaling results clearly demonstrate that slow resources, such as CPUs, can effectively contribute to parallel training and accelerate convergence.
Therefore, we conclude that the proposed biased local SGD enables effective utilization of heterogeneous system resources for parallel neural network training.


\subsection{Ablation Study}

\textbf{Evaluation of Impact of Bias} —
Table~\ref{tab:c2g8} compares unbiased and biased variants of local SGD.
We fix $\tau_F = 32$ and vary $\tau_S$ among 1, 4, and 16 to simulate different performance gaps between fast and slow resources.
The wall-clock time is fixed to the same value for each dataset.
We first observe that unbiased local SGD suffers a non-negligible drop in accuracy as $\tau_S$ decreases.
As the performance gap between compute resources widens, the accuracy achieved within a fixed time budget degrades more significantly.
This trend is also evident in the fine-tuning benchmarks.
Second, when both types of bias, loss-based data sampling and weighted model aggregation, are introduced into local SGD, accuracy is consistently restored across all benchmarks.
These results demonstrate that well-controlled bias can accelerate local SGD and yield higher accuracy even under strong resource heterogeneity.

\textbf{Additional Results} --
To further verify the performance of our system-aware biased local SGD, we provide additional results in the Appendix, including comparisons across sampling methods, feature-wise evaluation, and experiments under both 1-CPU and 1-GPU configurations.

\section {Conclusion}

In this study, we demonstrated the potential of utilizing heterogeneous system resources, such as CPUs and GPUs, for parallel neural network training.
Our extensive empirical study shows that unbalanced local SGD is a practical optimization strategy, particularly for modern GPU servers, as it eliminates costly synchronization overhead between fast and slow compute resources.
Moreover, by introducing well-controlled bias into data sampling and model aggregation, unbalanced local SGD can match the accuracy of conventional synchronous SGD, while incurring no synchronization cost.
Extending our study on biased local SGD to multi-node training remains important future work.

\bibliographystyle{unsrt}  
\bibliography{references}  
\clearpage
\appendix
\section {Appendix}

The appendix is structured as follows:
\begin{itemize}
\item Section~\ref{appendix:settings} summarizes experimental settings corresponding to all the experimental results reported in the main manuscript
\item Section~\ref{appendix:discussion} presents additional experimental results and analyses.
\item Section~\ref{appendix:theory} provides proofs of our proposed lemmas and theorem.

\end{itemize}

\subsection {Experimental Settings} \label{appendix:settings}
To ensure reproducibility and provide a comprehensive understanding of our experimental setup, all computations were performed on a dedicated GPU server designed to simulate heterogeneous computing environments.
Trainings run on a GPU server equipped with AMD EPYC 7543 CPU and two NVIDIA RTX 4090 GPUs.
The system contains Ubuntu 24.04.1 LTS, Python 3.12.7, and CUDA 12.4.
Our deep learning models are implemented and trained using PyTorch, with supporting libraries such as torchvision for computer vision tasks and transformers for natural language processing benchmarks.
To orchestrate parallel execution across the heterogeneous compute resources (CPU and GPUs), we employ mpi4py library that enables efficient inter-process communication and synchronization in our system-aware biased local SGD framework.

\noindent
\textbf{Experiments in 1-CPU and 1-GPU Setting} --
For all experiments, we used the SGD optimizer and applied the \textit{MultiStepLR} scheduler to all other datasets.
Both unbalanced local SGD (unbiased) and biased local SGD (proposed) were evaluated
under the same conditions as those in Table~\ref{tab:c1g1_setting}.

\noindent
\textbf{Experiments in 2-CPU and 8-GPU Setting} --
For all experiments, we used the SGD optimizer and applied the \textit{MultiStepLR} scheduler to all other datasets.
Both unbalanced local SGD (unbiased) and biased local SGD (proposed) were evaluated
under the same conditions as those in Table \ref{tab:c2g8_setting}.

\begin{table}[h]
\footnotesize
\setlength{\tabcolsep}{4pt}
\centering
\begin{minipage}{0.48\columnwidth}
\centering
\caption{Experiments in 1-CPU and 1-GPU setting.}
\begin{tabular}{lcccc} \toprule
\multirow{2}{*}{Dataset} & Batch & \multirow{2}{*}{Epochs} & Learning & Weight \\
&Size&& Rate & Decay \\ \midrule
\multirow{2}{*}{CIFAR-10}& &&&\\
\multirow{2}{*}{(ResNet20)}& 128 & 100& 0.3 & 0.0001\\ 
                           &  & & (60, 80 decay) & \\ \midrule
Oxford & & & \\
Flowers 102  & 32 & 300 & 0.1 & 0.0001 \\
(WRN16)                & &&(180, 240 decay)&\\\midrule
\multirow{2}{*}{CIFAR-100}& & & \\
\multirow{2}{*}{(WRN28)}  & 128 & 100 & 0.1 & 0.0005\\ 
                             & & & (60, 80 decay) \\ \midrule
Tiny  & & & \\
ImageNet & 128 & 10 & 0.0001 & 0.0005 \\ 
(ViT) &  & &  (7 decay)\\  \midrule
\multirow{2}{*}{AG News} & & & \\
\multirow{2}{*}{(DistilBERT)}& 128 &10  & 0.001& 0.0001\\ 
                             &  & &(7 decay) &  \\  
\bottomrule
\end{tabular}
\label{tab:c1g1_setting}
\end{minipage}
\hfill
\begin{minipage}{0.48\columnwidth}
\centering
\caption{Experiments in 2-CPU and 8-GPU setting.}
\begin{tabular}{lcccc} \toprule
\multirow{2}{*}{Dataset} & Batch & \multirow{2}{*}{Epochs} & Learning & Weight \\
&Size&& Rate & Decay \\ \midrule
\multirow{2}{*}{CIFAR-10}& &&&\\
\multirow{2}{*}{(ResNet20)}& 32 & 100 & 0.1 & 0.0001 \\
                           &  & & (60, 80 decay) & \\ \midrule
Oxford & & & \\
Flowers 102  & 11 & 300 & 0.1 & 0.0001 \\
(WRN16)                & &&(180, 240 decay)&\\\midrule
\multirow{2}{*}{CIFAR-100}& & & \\
\multirow{2}{*}{(WRN28)}  & 32 & 200 & 0.2 & 0.0005 \\
                             & & & (120, 160 decay)\\ \midrule
Tiny  & & & \\
ImageNet & 32 & 10 & 0.0001 & 0.0005 \\
(ViT) &  & & (7 decay)\\  \midrule
\multirow{2}{*}{AG News} & & & \\
\multirow{2}{*}{(DistilBERT)}& 32 & 10 & 0.001 & 0.0001\\
 &  & & (7 decay) &  \\  
\bottomrule
\end{tabular}
\label{tab:c2g8_setting}
\end{minipage}

\end{table}

\subsection {Additional Results and Analysis} \label{appendix:discussion}
\noindent
\textbf{Random Sampling Scheme Comparison} --
In the proposed system-aware biased Local SGD, CPU worker groups and GPU worker groups independently sample data.
This design enables each worker group to access the entire dataset without restrictions, potentially allowing some data to be sampled by both groups.
Under this setting, fast workers focus on learning from high-loss (challenging) data, while slow workers train on randomly sampled data.
Our empirical results well demonstrate that this separated sampling strategy leads to more effective local SGD training on heterogeneous system resources, achieving the highest accuracy among all possible random sampling schemes.

Table~\ref{tab:cpu_or_gpu} shows CIFAR-10, Oxford Flowers 102, and CIFAR-100 benchmark results obtained with three different random sampling schemes: unbiased uniform sampling, unified sampling, and our proposed separated sampling.
The \textit{Unified Sampling} indicates the setting where fast workers first sample the high-loss data and then slow workers train on the remaining data.
The \textit{Separated Sampling} is our proposed method described in the main manuscript.
As shown, our proposed \textit{Separated Sampling} achieves the best accuracy.

The \textit{Separated Sampling} has several benefits as follows.
First, by sampling data for slow workers independently of the fast workers, their gradient estimates remain unbiased, leading to more effective training.
Second, this design allows slow workers to still sample high-loss data.
As a result, these challenging samples may be used in more updates, introducing a bias toward harder examples.
Third, there is no synchronization points between slow and fast workers.
In \textit{Unified Sampling}, the fast workers should wait until all the slow workers finish the data sampling.
Thus, it causes additional blocking points increasing the synchronization cost.
For the above reasons, we argue that the proposed separated random sampling is the best design choice for efficient local SGD training.

\begin{table}[h]
\footnotesize
\centering
\caption{
    Performance comparison of data sampling methods (2-CPU and 8-GPU setting). The $\tau_S$ is set 1 and $\tau_F$ is fixed to 32.
}
\begin{tabular}{lcccc} \toprule
Dataset  & Data Allocation & Unified Sampling(Acc) & Separated Sampling(Acc) & Unbias(Acc) \\ 
\midrule
CIFAR-10 & High-loss → CPU & $88.79\pm 0.3\%$ &$89.05\pm 0.3\%$ &\multirow{2}{*}{$87.98\pm 0.2\%$} \\ 
(ResNet20)& High-loss → GPU & $88.96\pm 0.1\%$& $\mathbf{89.16}\pm0.1 \%$ & \\ 
\midrule

Oxford Flowers 102 & High-loss → CPU
&$71.57\pm 0.2\%$ & $70.20 \pm0.6 \%$ &\multirow{2}{*}{$66.08\pm 1.0\%$}  \\ 
(WRN16)& High-loss → GPU
& $72.30\pm 0.2\%$ &$\mathbf{73.09}\pm 1.4\%$ \\ 

  \midrule
CIFAR-100 & High-loss → CPU
& $77.54\pm 0.3\%$ & $77.61 \pm 0.3\%$ &\multirow{2}{*}{$77.01\pm 0.2 \%$}   \\ 
(WRN28)& High-loss → GPU 
& $77.74\pm 0.1\%$ & $\mathbf{77.97} \pm0.4 \%$ \\  \midrule
\end{tabular}
\label{tab:cpu_or_gpu}
\vspace{1em}
\end{table}

\begin{figure}[h]
\centering
\includegraphics[width=1\columnwidth]{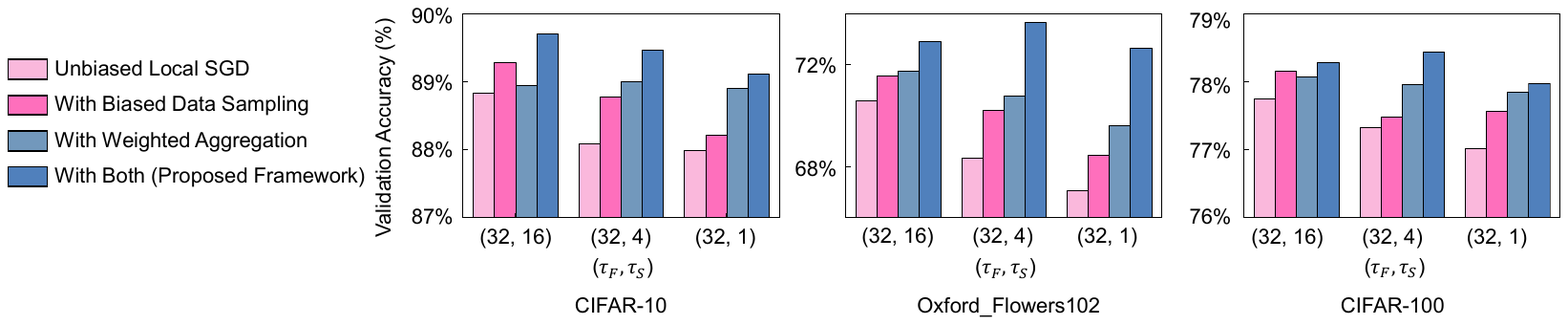}
\vspace{-1.3em}
\caption{
    Feature-wise accuracy comparison with various combinations of fast and slow resources.
}
\label{fig:bar}
\vspace{1em}
\end{figure}

\textbf{Feature-wise Evaluation} --
Figure~\ref{fig:bar} presents a performance comparison across four bias settings:
(1) unbiased local SGD, (2) local SGD with biased data sampling, (3) local SGD with weighted aggregation, and (4) local SGD with both techniques.
We fix $\tau_F = 32$ and vary $\tau_S$ among 1, 4, and 16 to simulate different performance gaps between fast and slow compute resources.
A smaller $\tau_S$ indicates a slower CPU (e.g., $\tau_F=32, \tau_S=16$ implies a CPU twice as slow as the GPU).
As $\tau_S$ decreases, unbiased local SGD exhibits a sharp accuracy drop across all benchmarks.
This degradation primarily arises from the reduced number of updates performed by the slower workers.
Introducing bias through data sampling effectively restores accuracy, and weighted model aggregation similarly improves accuracy across all benchmarks.
Finally, combining both techniques yields a substantial accuracy improvement.

\textbf{Effect of Bias in 1-CPU and 1-GPU Setting} -- We present accuracy comparisons between unbiased and biased local SGD in Table~\ref{tab:c1g1}.
Here, we consider a more practical setting in which only one CPU and one GPU are available.
We fix $\tau_F = 32$ and vary $\tau_S$ among 1, 4, and 16 to simulate different performance gaps between fast and slow resources.
For each dataset, the wall-clock time is fixed to the same budget.
We observe similar results to those shown in Table~\ref{tab:c2g8}.
The biased local SGD consistently achieves higher accuracy across all benchmarks, regardless of the performance gap between the CPU and GPU.
These results demonstrate that a well-controlled bias can accelerate local SGD and yield higher accuracy even under severe resource heterogeneity.

\clearpage

\begin{table}[h]
\caption{
    [\textbf{1-CPU and 1-GPU}] Performance comparison between the unbiased local SGD and the proposed biased local SGD.
}\label{tab:c1g1}
\footnotesize
\centering
\begin{tabular}{llc>{\columncolor{gray!10}}c} \toprule
\multirow{2}{*}{Dataset} & \multirow{2}{*}{($\tau_F$, $\tau_S$)} & \multirow{2}{*}{Unbiased} & Biased \\
&&& (proposed) \\ \midrule
\multirow{2}{*}{CIFAR-10}& (32,16) & $91.16\pm 0.1\%$ & $\mathbf{91.63} \pm0.2 \%$  \\ 
\multirow{2}{*}{(ResNet20)}& (32,4) & $90.58\pm0.1 \%$ & $\mathbf{91.22} \pm0.2 \%$  \\ 
& (32,1) & $90.21\pm0.3 \%$ & $\mathbf{91.08} \pm0.1 \%$  \\  \midrule
Oxford & (32,16) & $77.70\pm 0.9\%$ & $\mathbf{81.62} \pm0.1 \%$  \\ 
Flowers 102 & (32,4) & $73.38\pm0.1 \%$ & $\mathbf{78.63} \pm 0.3 \%$  \\  
(WRN16) & (32,1) & $72.01\pm 0.6\%$ & $\mathbf{78.58} \pm0.9 \%$  \\   \midrule
\multirow{2}{*}{CIFAR-100}& (32,16) & $77.59\pm0.4 \%$ & $\mathbf{78.82} \pm0.4 \%$  \\ 
\multirow{2}{*}{(WRN28)} & (32,4) & $77.32\pm 0.1\%$ & $\mathbf{78.01} \pm0.2 \%$  \\ 
& (32,1) & $76.49\pm0.2 \%$ & $\mathbf{77.98} \pm0.1 \%$  \\  \midrule
Tiny & (32,16) & $90.09\pm 0.1\%$ & $\mathbf{90.78} \pm0.1 \%$  \\ 
ImageNet & (32,4) & $89.45\pm0.1 \%$ & $\mathbf{90.69} \pm0.1 \%$  \\ 
(ViT) & (32,1) & $88.73\pm 0.2\%$ & $\mathbf{90.48} \pm 0.2 \%$  \\   \midrule
\multirow{2}{*}{AG News} & (32,16) & $93.71\pm 0.1\%$ & $\mathbf{94.08} \pm0.1 \%$  \\ 
\multirow{2}{*}{(DistilBERT)} & (32,4) & $93.13\pm0.1 \%$ & $\mathbf{93.97} \pm0.1 \%$  \\ 
& (32,1) & $92.98\pm0.1 \%$ & $\mathbf{93.95} \pm0.1 \%$  \\  \bottomrule
\end{tabular}
\end{table}


\subsection{Theoretical Analysis} \label{appendix:theory}
Herein, we provide proof of Theorem and Lemma shown in Section~\ref{sec:theory}.
We consider non-convex and smooth optimization problems as follows.
\begin{align}
    \min_{x \in \mathbb{R}^d} F(x) := \frac{1}{n} \sum_{i=1}^n f(x, \xi_i), \label{eq:problem}
\end{align}
where $f(x)$ is a loss function, $x \in \mathbb{R}^d$ is the $d$-dimensional model parameter vector, $n$ is the number of training samples, and $\xi_i$ is the $i^{th}$ training sample.







We now analyze the convergence properties of biased local SGD in IID settings.
The assumptions can be found in Section~\ref{sec:theory}.

\begin{lemma}\label{lemma:framework}
(framework) Under assumption 1 $\sim$ 3, if the learning rate $\eta \leq \frac{1}{\mathcal{L}\tau}$, we have
\begin{align}
\mathbb{E}_p\left[ F(x_{t+1}) \right] & \leq \mathbb{E}_p\left[ F(x_t) \right] -\frac{\eta}{2} \mathbb{E}_p\left[ \left\| \nabla F(x_t) \right\|^2 \right] + \frac{\mathcal{L}\eta^2 \tau^2}{2} \sigma_p^2. \label{eq:lemma1}
\end{align}
\end{lemma}

\begin{proof}
Based on Assumption 1, we begin with:
\begin{align}
    \mathbb{E}_p\left[ F(x_{t+1}) \right] &\leq \mathbb{E}_p\left[ F(x_t) \right] + \mathbb{E}_p\left[ \langle \nabla F(x_t), x_{t+1} - x_t \rangle \right] + \frac{\mathcal{L}}{2} \mathbb{E}_p\left[ \left\| x_{t+1} - x_t \right\|^2 \right] \nonumber \\
    &= \mathbb{E}_p\left[ F(x_t) \right] -\eta \mathbb{E}_p\left[ \langle \nabla F(x_t), \bar\Delta_t \rangle \right] + \frac{\mathcal{L}\eta^2}{2} \mathbb{E}_p\left[ \left\| \bar\Delta_t \right\|^2 \right] \nonumber \\
    &= \mathbb{E}_p\left[ F(x_t) \right] -\eta \mathbb{E}_p\left[ \langle \nabla F(x_t), \bar\Delta_t \rangle \right] \nonumber \\
    &\quad + \frac{\mathcal{L}\eta^2}{2} \mathbb{E}_p\left[ \left\| \bar\Delta_t - \sum_{i=1}^{m} \sum_{j=0}^{\tau - 1} w_i \nabla F_p(x_{t,j}^i) \right\|^2 \right] + \frac{\mathcal{L}\eta^2}{2} \mathbb{E}_p\left[ \left\| \sum_{i=1}^{m} \sum_{j=0}^{\tau - 1} w_i \nabla F_p(x_{t,j}^i) \right\|^2 \right] \nonumber
\end{align}
Then, based on Assumption 3, we have:
\begin{align}
    \mathbb{E}_p \left[ F(x_{t+1})\right] &\leq \mathbb{E}_p\left[ F(x_t) \right] -\eta \mathbb{E}_p\left[ \langle \nabla F(x_t), \bar\Delta_t \rangle \right] + \frac{\mathcal{L}\eta^2\tau^2}{2}\sigma_p^2 + \frac{\mathcal{L}\eta^2}{2} \mathbb{E}_p\left[ \left\| \sum_{i=1}^{m} \sum_{j=0}^{\tau - 1} w_i \nabla F_p(x_{t,j}^i) \right\|^2 \right] \nonumber \\
    &= \mathbb{E}_p\left[ F(x_t) \right] -\eta \mathbb{E}_p\left[ \langle \nabla F(x_t), \bar\Delta_t \rangle \right] + \frac{\mathcal{L}\eta^2\tau^2}{2}\sigma_p^2  \nonumber \\
    &\quad + \frac{\mathcal{L}\eta^2}{2} \mathbb{E}_p\left[ \left\| \sum_{i=1}^{m} \sum_{j=0}^{\tau - 1} w_i \nabla F_p(x_{t,j}^i) -\tau \nabla F(x_{t}) \right\|^2 \right] - \frac{\mathcal{L}\eta^2}{2} \mathbb{E}_p\left[ \left\| \tau \nabla F(x_t) \right\|^2 \right] \nonumber \\
    &\quad + \mathcal{L}\eta^2 \mathbb{E}_p \left[ \langle \sum_{i=1}^{m}\sum_{j=0}^{\tau - 1} w_i \nabla F_p(x_{t,j}^{i}), \tau \nabla F(x_t) \rangle \right] \nonumber
\end{align}
By rearranging the terms, we have:
\begin{align}
    \mathbb{E}_p\left[ F(x_{t+1}) \right] &\leq \mathbb{E}_p\left[ F(x_t) \right] -\eta\left(1 - \mathcal{L}\eta \tau \right) \underset{T}{\underbrace{ \mathbb{E}_p\left[ \langle \nabla F(x_t), \bar\Delta_t \rangle \right] }} + \frac{\mathcal{L}\eta^2\tau^2}{2}\sigma_p^2 - \frac{\mathcal{L}\eta^2 \tau^2}{2} \mathbb{E}_p \left[ \left\| \nabla F(x_t) \right\|^2 \right] \label{eq:T} \\
    &+ \frac{\mathcal{L}\eta^2}{2} \mathbb{E}_p \left[ \left\| \sum_{i=1}^{m}\sum_{j=0}^{\tau - 1} w_i \nabla F_p(x_{t,j}^i) - \tau \nabla F(x_t) \right\|^2 \right]. \nonumber 
\end{align}

Here, $T$ is bounded as follows.
\begin{align}
    T &= \mathbb{E}_p\left[ \langle \nabla F(x_t), \bar\Delta_t \rangle \right] \nonumber \\
    &= \mathbb{E}_p\left[ \langle \nabla F(x_t), \sum_{i=1}^{m} \sum_{j=0}^{\tau-1} w_i \nabla F_p(x_{t,j}^{i}) \rangle \right] \nonumber \\
    & = \mathbb{E}_p \left[ \langle \nabla F(x_t), \sum_{i=1}^{m} \sum_{j=0}^{\tau-1} w_i \nabla F_p(x_{t,j}^{i}) - \tau \nabla F (x_t) \rangle \right] + \tau \mathbb{E}_p\left[ \left\| \nabla F(x_t) \right\|^2 \right] \nonumber \\
    & \geq \left(\tau -\frac{1}{2}\right) \mathbb{E}_p \left[ \left\| \nabla F(x_t) \right\|^2 \right] -\frac{1}{2} \mathbb{E}_p \left[ \left\| \sum_{i=1}^{m} \sum_{j=0}^{\tau-1} w_i \nabla F_p(x_{t,j}^{i}) - \tau \nabla F(x_t) \right\|^2 \right], \label{eq:T_bound}
\end{align}
where (\ref{eq:T_bound}) holds since $\langle a, b \rangle \geq -\frac{1}{2}\| a \|^2 - \frac{1}{2} \| b \|^2$.

Then, by plugging (\ref{eq:T_bound}) into (\ref{eq:T}), we have:
\begin{align}
    \mathbb{E}_p\left[ F(x_{t+1}) \right] & \leq \mathbb{E}_p\left[ F(x_t) \right] + \frac{\mathcal{L}\eta^2 \tau^2}{2} \sigma_p^2 \nonumber \\ 
    &\quad - \left( \frac{\mathcal{L}\eta^2\tau^2}{2} + \eta(1 - \mathcal{L}\eta\tau)\left(\tau - \frac{1}{2}\right) \right) \mathbb{E}_p\left[ \left\| \nabla F(x_t) \right\|^2 \right] \nonumber \\
    &\quad + \left(\frac{\mathcal{L}\eta^2}{2} + \frac{\eta(1 - \mathcal{L}\eta \tau)}{2} \right) \mathbb{E}_p \left[ \left\| \sum_{i=1}^{m}\sum_{j=0}^{\tau - 1} w_i \nabla F_p(x_{t,j}^i) - \tau \nabla F(x_t) \right\|^2 \right] \label{eq:lr0} \\
    &\leq \mathbb{E}_p\left[ F(x_t) \right] + \frac{\mathcal{L}\eta^2 \tau^2}{2} \sigma_p^2 \label{eq:lr1} \\
    &\quad - \left( \frac{\mathcal{L}\eta^2\tau^2}{2} + \eta(1 - \mathcal{L}\eta\tau)\left(\tau - \frac{1}{2}\right) \right) \mathbb{E}_p\left[ \left\| \nabla F(x_t) \right\|^2 \right] \nonumber
\end{align}

where (\ref{eq:lr1}) holds because the final term on the right-hand side in (\ref{eq:lr0}) is negative if $\eta \leq \frac{1}{\mathcal{L}(\tau - 1)}$.

Finally, (\ref{eq:lr1}) can be simplified if $\eta \leq \frac{1}{\mathcal{L}\tau}$ as follows.
\begin{align}
    \mathbb{E}_p\left[ F(x_{t+1}) \right] & \leq \mathbb{E}_p\left[ F(x_t) \right] + \frac{\mathcal{L}\eta^2 \tau^2}{2} \sigma_p^2 \nonumber \\
    &\qquad - \left( \frac{\mathcal{L}\eta^2\tau^2}{2} + \eta(1 - \mathcal{L}\eta\tau)\left(\tau - \frac{1}{2}\right) \right) \mathbb{E}_p\left[ \left\| \nabla F(x_t) \right\|^2 \right] \nonumber \\
    &\leq \mathbb{E}_p\left[ F(x_t) \right] - \frac{\eta}{2} \mathbb{E}_p\left[ \left\| \nabla F(x_t) \right\|^2 \right] + \frac{\mathcal{L}\eta^2 \tau^2}{2} \sigma_p^2. \label{eq:lr2}
\end{align}
If $\eta \leq \frac{1}{\mathcal{L}\tau}$ and $\tau \geq 1$, (\ref{eq:lr2}) holds, and thus (\ref{eq:lr3}) is always true.
\begin{align}
    0 & \leq \frac{\mathcal{L}\eta^2 \tau^2}{2} + \eta(1 - \mathcal{L}\eta\tau)(\tau - \frac{1}{2}) - \frac{\eta}{2} \nonumber \\
    & \leq \frac{\mathcal{L}\eta^2 \tau^2}{2} + \eta\tau - \frac{\eta}{2} \nonumber \\
    & \leq \eta\tau \left( \frac{\mathcal{L}\eta\tau}{2} + 1\right)  - \frac{\eta}{2} \leq \eta \left( 2\tau - \frac{1}{2}\right). \label{eq:lr3}
\end{align}
\end{proof}


\begin{theorem}
    Under assumption $1 \sim 3$, if the learning rate $\eta \leq \frac{1}{\mathcal{L}\tau}$ and $\tau \geq 1$, the average gradient norm becomes
\begin{align}
    \frac{1}{T}\sum_{t=1}^{T} \mathbb{E}_p \left[ \left\| \nabla F(x_{t}) \right\|^2 \right] &\leq \frac{2}{\eta T} \left( \mathbb{E}_p \left[ F(x_0) \right] - \mathbb{E}_p \left[ F(x_{T}) \right] \right) \nonumber \\
    &\qquad + \mathcal{L}\eta \tau^2 \sigma_p^2.
\end{align}
\end{theorem}

\begin{proof}
By rearranging terms and averaging (\ref{eq:lemma1}) across $T$ communication rounds, we have a telescope sum as follows.
\begin{align}
    &\frac{\eta}{2T}\sum_{t=1}^{T} \mathbb{E}_p \left[ \left\| \nabla F(x_{t}) \right\|^2 \right] \nonumber \\
    &\quad \leq \frac{1}{T} \sum_{t=0}^{T-1}\left( \mathbb{E}_p \left[ F(x_t) \right] - \mathbb{E}_p \left[ F(x_{t+1}) \right] \right) + \frac{1}{T}\sum_{t=0}^T \frac{\mathcal{L}\eta^2 \tau^2}{2} \sigma_p^2 \nonumber\\
    &\quad \leq \frac{1}{T} \left( \mathbb{E}_p \left[ F(x_0) \right] - \mathbb{E}_p\left[ F(x_{T}) \right] \right) + \frac{\mathcal{L}\eta^2 \tau^2}{2} \sigma_p^2. \nonumber
\end{align}
Finally, dividing both sides by $\frac{\eta}{2}$ yields,
\begin{align}
    \frac{1}{T}\sum_{t=1}^{T} \mathbb{E}_p\left[ \left\| \nabla F(x_{t}) \right\|^2 \right] &\leq \frac{2}{\eta T} \left( \mathbb{E}_p \left[ F(x_0) \right] - \mathbb{E}_p \left[ F(x_{T}) \right] \right) + \mathcal{L}\eta \tau^2 \sigma_p^2. \nonumber
\end{align}
\end{proof}

\clearpage

\end{document}